\def\paperlanguage{}
\pdfoutput=1

\documentclass[letterpaper, 10 pt, conference]{ieeeconf}  

\usepackage{bm}
\usepackage{cite}
\usepackage{flushend}
\include{preamble}

\newcommand{\ctext}[1]{\raise0.2ex\hbox{\textcircled{\scriptsize{#1}}}}

\IEEEoverridecommandlockouts                              

\title{\LARGE \textbf
  {
    \switchlanguage%
    {%
      VQA-based Robotic State Recognition Optimized with\\Genetic Algorithm
    }%
    {%
      遺伝的アルゴリズムにより最適化されたVQAによるロボットの状態推定
    }%
  }
}

\author{Kento Kawaharazuka$^{1}$, Yoshiki Obinata$^{1}$, Naoaki Kanazawa$^{1}$, Kei Okada$^{1}$, and Masayuki Inaba$^{1}$
  \thanks{$^{1}$ The authors are with the Department of Mechano-Informatics, Graduate School of Information Science and Technology, The University of Tokyo, 7-3-1 Hongo, Bunkyo-ku, Tokyo, 113-8656, Japan.
    {\texttt\small [kawaharazuka, obinata, kanazawa, k-okada, inaba]@jsk.t.u-tokyo.ac.jp}
  }
}
\begin{document}

\maketitle
\thispagestyle{empty}
\pagestyle{empty}

\begin{abstract}
  \switchlanguage%
  {%
    State recognition of objects and environment in robots has been conducted in various ways.
    In most cases, this is executed by processing point clouds, learning images with annotations, and using specialized sensors.
    In contrast, in this study, we propose a state recognition method that applies Visual Question Answering (VQA) in a Pre-Trained Vision-Language Model (PTVLM) trained from a large-scale dataset.
    By using VQA, it is possible to intuitively describe robotic state recognition in the spoken language.
    On the other hand, there are various possible ways to ask about the same event, and the performance of state recognition differs depending on the question.
    Therefore, in order to improve the performance of state recognition using VQA, we search for an appropriate combination of questions using a genetic algorithm.
    We show that our system can recognize not only the open/closed of a refrigerator door and the on/off of a display, but also the open/closed of a transparent door and the state of water, which have been difficult to recognize.
  }%
  {%
    これまで, ロボットにおける物体や環境の状態認識は様々な方法で行われてきた.
    大抵の場合, これらはポイントクラウドの処理やアノテーションを使った画像の学習, 専用のセンサを使って行われる.
    これに対して本研究では, 大規模なデータセットから学習された視覚-言語モデルを用いたVisual Question Answeringを応用する状態認識手法を行う.
    VQAを用いることで, 言語による直感的な状態認識記述が可能である.
    一方, 同じ事象に対しても, その質問は様々な方法が考えられ, その質問ごとに状態認識の性能は異なる.
    そこで本研究では, VQAを用いた状態認識の性能を向上させるため, 遺伝的アルゴリズムによる適切な質問文の組み合わせの探索を行う.
    これにより, 冷蔵庫ドアの開閉やディスプレイのオンオフだけでなく, 認識の難しい透明なドアの開閉や水の状態認識が, 高い精度で可能であることを示す.
  }%
\end{abstract}

\section{INTRODUCTION}\label{sec:introduction}
\switchlanguage%
{%
  When a robot moves through space and performs a task, recognition of surrounding objects, tools, and environment is indispensable.
  For example, the robot needs to recognize the open/closed state of doors to rooms and elevators, the on/off of lights, etc. when moving around \cite{saito2011subwaydemo, okada2006tool}.
  In addition, it is necessary to recognize the open/closed state of refrigerator and cabinet doors, the on/off of displays, whether water is left running from the faucet, etc. when patrolling the area.
  So far, these state recognitions have been mainly based on human programming to extract features from raw images or point clouds \cite{chin1986recognition, borgsen2014door}, human annotation of images and training with neural networks \cite{li2020modifiedyolov3}, or the attachment of appropriate sensors depending on the state to be recognized \cite{takahata2020coaxial}.
  In other words, we have constructed a recognizer for each state that we want to recognize.
  On the other hand, with these methods, the number of recognizers and the data to be collected increases as the number of states to be recognized increases, which requires a large amount of time for system construction and makes resource management difficult.
  In addition, when humans walk around, they are constantly recognizing various states that cannot be easily programmed.
  These states include the open/closed of transparent doors, the state of water, etc.
  These states are not so easy to recognize for robots, and various studies have been conducted on the recognition alone \cite{khaing2011transparent}.
}%
{%
  ロボットが空間を移動しタスクをこなす際に, 周囲の物体や道具, 環境の認識は不可欠である.
  例えば, 移動の際に部屋やエレベータドアの開閉, 明かりのオンオフ等を認識する必要がある\cite{saito2011subwaydemo, okada2006tool}.
  また, 見回りの際に, 冷蔵庫や棚のドアの開閉, ディスプレイのオンオフ, キッチンの水の出しっぱなし等も認識する必要がある.
  これまで, それらの認識器は, 主に人のプログラミングによって生画像やポイントクラウドから何らかの処理を行って特徴量を抽出する\cite{chin1986recognition, borgsen2014door}, 画像をアノテーションしニューラルネットワークで学習することで行う\cite{li2020modifiedyolov3}, 認識したい状態に応じて適切なセンサを取り付けるといった方法\cite{takahata2020coaxial}で行われてきた.
  つまり, 認識したい状態ごとに人が認識器を構築していくという方法である.
  一方, この方法では, 認識したい状態が増えるほど認識器の数や収集しなければならないデータも増え, システムの構築に多大な時間を要し, かつリソース管理も難しくなる.
  また, 人間が普段歩き回る際には, 簡単にはプログラミングできないような様々な状態認識を常に行っている.
  それは, 透明なドアの開閉や水の認識などを含み, これらはそう簡単ではなく, その認識単体でも様々な研究が行われてきている\cite{khaing2011transparent}.
}%

\begin{figure}[t]
  \centering
  \includegraphics[width=0.95\columnwidth]{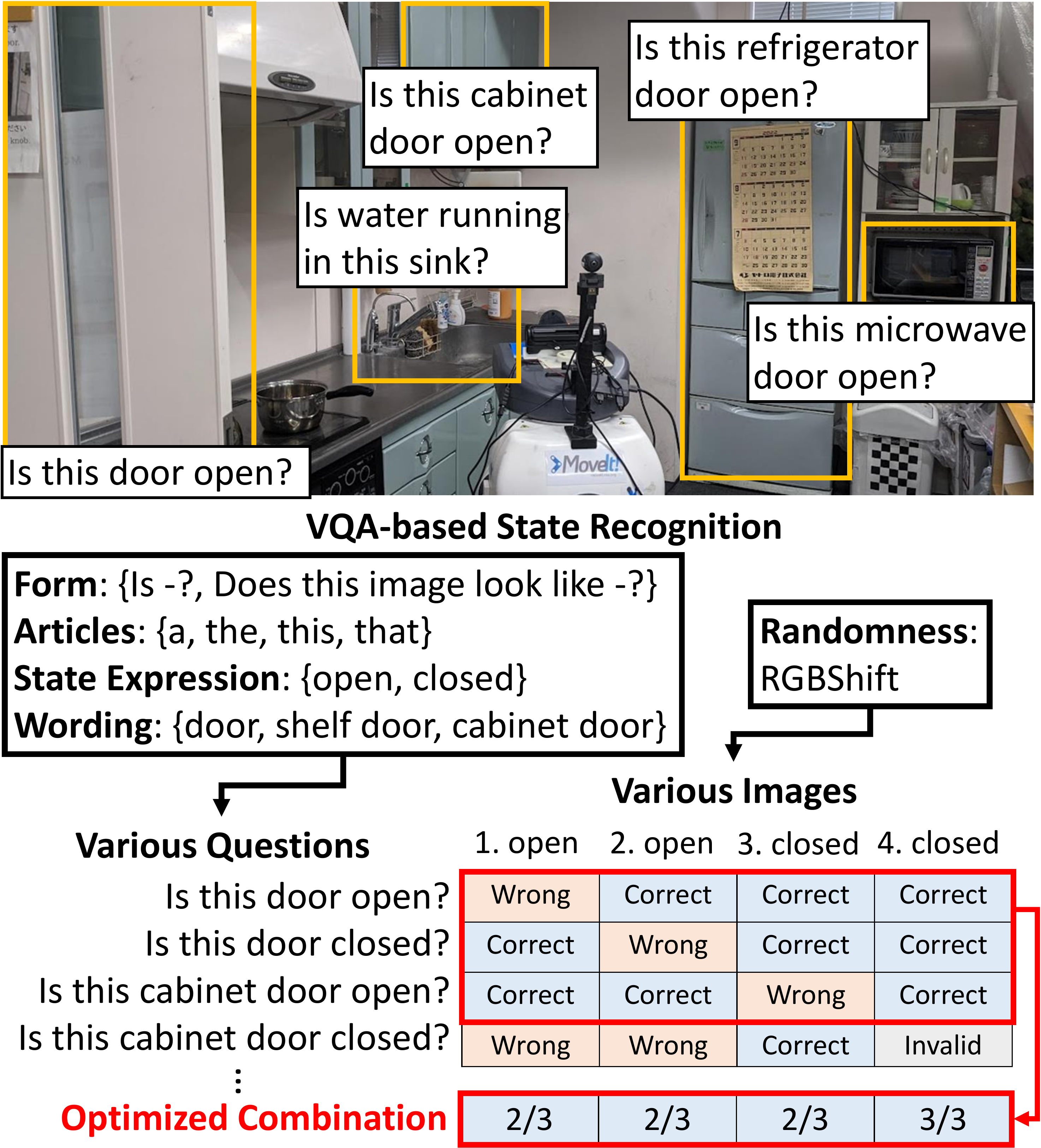}
  \vspace{-0.6ex}
  \caption{The concept of this study. The combination of appropriate questions is optimized with genetic algorithm for binary state recognition using VQA.}
  \label{figure:concept}
  \vspace{-1.8ex}
\end{figure}

\switchlanguage%
{%
  Therefore, in this study, we propose a method of state recognition using the spoken language.
  We apply Visual Question Answering (VQA) \cite{antol2015vqa} in a Pre-Trained Vision-Language Model (PTVLM) \cite{radford2021clip, li2022largemodels} trained from a large-scale dataset.
  The state recognition is performed by asking a question about the current image and getting the answer in sentence form.
  By using the spoken language, it is possible to recognize even states that cannot be easily described by a program.
  On the other hand, preliminary experiments have shown that the recognition performance depends greatly on the question, especially on the form, articles, state expressions, and wording that are used.
  Since the states that are easy to recognize and the ones that are hard to recognize are different for each question, the states can only be partially recognized accurately with a single question.
  Therefore, in this study, we develop a method to search for appropriate question combinations using a genetic algorithm and construct a state recognizer with high performance.
  We show that our system can perform the state recognition of not only the open/closed of a refrigerator door and the on/off of a display, but also the open/closed of a transparent door and the state of water, which have been difficult to recognize.
  This study shows that an appropriate recognizer with high performance can be automatically generated by simply providing candidate questions in VQA, and that state recognition can be performed simply and quickly using the spoken language and a single vision-language model, thus dramatically improving the recognition behavior of robots.
}%
{%
  そこで本研究では, 言語を用いた状態認識器の構築方法を提案する.
  現在盛んになってきた, 大規模なデータセットを用いて学習された視覚-言語モデル\cite{radford2021clip, li2022largemodels}における, Visual Question Answering (VQA) \cite{antol2015vqa}を応用する.
  現在の画像に対して言語で質問をし, その回答を言語で得ることで状態認識を行う.
  言語を用いることで, 簡単にはプログラムで記述できないような状態さえも認識できるようになる.
  一方で, その認識性能は質問の文章, 特に質問の形式や冠詞, 状態表現や言葉遣い等に大きく依存することが予備実験から分かってきた.
  それぞれの質問にとって認識しやすい状態と認識しにくい状態が異なるため, 単一の質問では一部の状態しか正確に認識することができない.
  そこで, 適切な質問の組み合わせを発見することで, どの状態も正しく認識可能な認識器を構築できることに気がついた.
  本研究では, 適切な質問の組み合わせを遺伝的アルゴリズムを用いて探索し, 性能の高い状態認識器を構築する手法を開発する.
  構築した認識器が, 冷蔵庫ドアの開閉やディスプレイのオンオフだけでなく, 認識の難しい透明なドアの開閉や水の状態認識も可能であることを示す.
  本研究により, VQAにおける質問の候補を与えるだけで, 自動的に性能の高い適切な認識器が生成可能であり, 言語を使い簡易かつ素早く, 単一のモデルで状態認識が可能となり, ロボットの認識行動性能を飛躍的に向上可能であることを示す.
}%

\section{VQA-based Robotic State Recognition Optimized with Genetic Algorithm} \label{sec:proposed}

\subsection{VQA-based Robotic State Recognition Using Pre-Trained Vision-Language Models} \label{subsec:recognition}
\switchlanguage%
{%
  First, we describe a state recognition method using VQA.
  VQA is a task to obtain an answer $A$ by asking a question $Q$ to an image $V$.
  In particular, we deal with binary state recognition, which can be easily used for action branching of robots.
  In other words, the task is to get a Yes or No answer by asking a question such as ``Is -?'' or ``Does -?''.
  On the other hand, such questions do not always result in Yes or No answers, but often in other phrases.
  Therefore, the obtained answers are classified into three categories: Correct, Wrong, and Invalid.
  Here, by augmenting the image or by changing the question, multiple answers can be obtained for the same image.
  The number of Correct, Wrong, and Invalid responses obtained is expressed as $N_{\{correct, wrong, invalid\}}$, and we define Correct Rate $R_{correct}$ as $N_{correct}/(N_{correct}+N_{wrong})$ and Invalid Rate $R_{invalid}$ as $N_{nvalid}/(N_{correct}+N_{wrong}+N_{invalid})$ ($N_{invalid}$ is not included in the denominator in $R_{correct}$).
  If $R_{correct}$ is greater than 0.5 for a certain image, then the image is correctly recognized.

  Next, we describe how to augment images and generate various questions regarding the image.
  First, we use RGBShift as image augmentation, which adds a random value sampled from a uniform distribution in [-0.1, 0.1] to each RGB value.
  This allows us to obtain multiple responses to the same image and question and to take their average (in this study, we prepared six augmented images for each image-question pair).
  Of course, other augmentation such as Gaussian noise are also possible.
  Next, we generate various types of questions by combining four types of sentence variation: question form, article, state expression, and wording.
  The form refers to the way of asking a question, and in this study, the forms of ``Is -?'' such as ``Is this door open?'', and ``Does -?'' such as ``Does this image look like this door is open?'' are used.
  The article refers to the use of {``a'', ``the'', ``this'', ``that''} for the target object.
  The state expression refers to the use of antonyms, for example, ``closed'' for ``open'', ``is not'' for ``is'', and so on.
  The wording refers to, for example, the use of ``display'', ``monitor'', ``TV'', etc. for a monitor, and in this study, up to four types of wording are prepared for each state recognizer.
  Therefore, we will prepare a maximum of $2\times4\times2\times4$, that is, 64 questions.
  Here, there are at most $2^{64}-1$, or about $10^{19}$ combinations of questions.
  Of course, the number of question types is not limited to these, but can be increased in any number of ways, such as singular and plural, etc.

  In this study, we use OFA \cite{wang2022ofa} as a Pre-Trained Vision-Language Model that can be used for VQA.
  OFA is a model with high generalization ability by learning multiple tasks such as image captioning, visual grounding, and text-to-image generation at the same time.
}%
{%
  まず, VQAを使った状態認識手法について述べる.
  VQAは画像Vに対して質問Qを与えることで, 回答Aを得るタスクである.
  これを応用することで状態認識を行う.
  そして本研究では特に, ロボットの行動分岐等に容易に用いることが可能な, 二値状態認識を取り扱う.
  つまり, ``Is -?''や``Does -?''等の質問を行うことで, YesまたはNoの回答を得ることが可能な問題設定である.
  一方で, ``Is -?''や``Does -?''の質問をしても, 常にYesまたはNoの回答が得られるとは限らず, それ以外の文章を返していまうことも多々ある.
  そのため, 得られた回答は, Correct, Wrong, Invalidの3つに分類される.
  この時, 画像にノイズを加えたり, その文章を変化させることで, 同じ画像に対して複数の回答を得ることができる.
  この際に得られたCorrect, Wrong, Invalidな回答の個数を, $N_{\{correct, wrong, invalid\}}$と表現し, 正解率$R_{correct}$を$N_{correct}/(N_{correct}+N_{wrong})$, 無効率$R_{invalid}$を$N_{nvalid}/(N_{correct}+N_{wrong}+N_{invalid})$と定義する(なお, 正解率において$N_{invalid}$は分母に含めない).
  ある画像に対して, $R_{correct}$が0.5を超えていれば, 正しい状態認識ができている.

  次に, 画像へのノイズや文章の変化の例を述べる.
  まず, 本研究では画像へのノイズとして, RGBそれぞれの値に対して, [-0.1, 0.1]の一様分布からサンプリングされたランダムな値を足し込む, RGBShiftを用いた.
  これによって, 同じ画像と質問でも複数の回答を得て, それらの平均を取ることができるようになる(本研究では一つの画像と質問のペアに対して, 6つのノイズを加えた画像を用意した).
  なお, もちろんその他ガウシアンノイズ等も可能である.
  次に, 文章の変化として, 文章の形式・冠詞・状態表現・言葉遣いという4つの変化の組み合わせによって, 多様な質問文を生成する.
  形式とは質問の仕方であり, 本研究では, ``Is this door open?''のような``Is -?''の形と, ``Does this image look like this door is open?''のような``Does -?''の形の2種類を用いる.
  冠詞とは, 対象となる物体に対して, {``a'', ``the'', ``this'' ``that''}の4種類を用いることである.
  状態表現とは, 対義語の使用を意味し, 例えば``open''に対して``closed'', ``is''に対して``is not''を用いることで2種類の表現を用意する.
  言葉遣いとは, 例えばモニターに対して, ``display''や``monitor'', ``TV''等を用いることを指し, 状態認識器ごとに1種類から4種類を用意する.
  つまり, 本研究では最大で$2\times4\times2\times4$の, 64の質問を用意することになる.
  この場合, 質問の組み合わせ方は最大で$2^{64}-1$, 約$10^{19}$存在する.
  もちろん質問の種類はこれだけではなく, 単数形や複数形, ここに示していないような質問形式など, その種類は如何用にも増やすことが可能である.

  なお, 本研究では, 大規模なデータセットにより訓練されたVQAが可能な事前学習済みの視覚-言語モデルとして, OFAを用いる\cite{wang2022ofa}.
  これは, Image CaptioningやVisual Grounding, Text-to-Image Generation等複数のタスクを同時に学習させることで, 高い汎化能力を得たモデルである.
}%

\begin{figure*}[t]
  \centering
  \includegraphics[width=1.9\columnwidth]{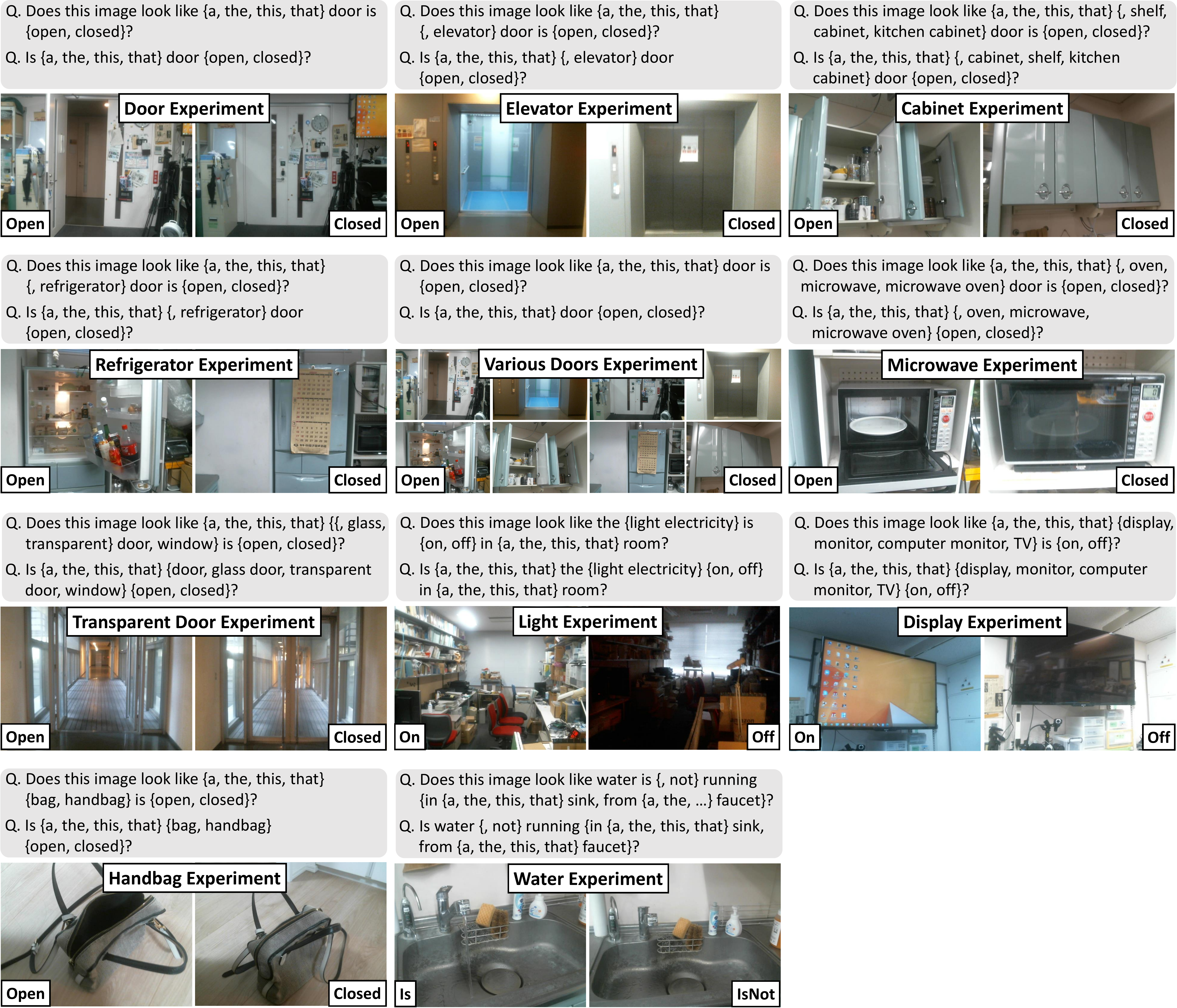}
  \vspace{-0.6ex}
  \caption{The set of questions and representative images for the door, elevator, cabinet, refrigerator, various doors, microwave, transparent door, light, display, handbag, and water experiment.}
  \label{figure:experiment}
  \vspace{-1.8ex}
\end{figure*}

\subsection{Optimization of Appropriate Question Combination Using Genetic Algorithm} \label{subsec:optimization}
\switchlanguage%
{%
  For the large number of questions mentioned in \secref{subsec:recognition}, we search for the combination of questions that can achieve the best accuracy.
  In this study, a genetic algorithm is used for this purpose.
  The question combination is represented by a binary vector $\bm{s}$ ($\in \{0, 1\}^{N_q}$, where $N_q$ denotes the number of all questions) and is optimized.
  Questions with the value of 1 are used, while questions with the value of 0 are not used.
  As a dataset for optimization, we collect $N^{train}_{data}=20$ images.
  For example, in the case of the open/closed state recognition of doors, we prepare 10 images of the door in the open state and 10 images of the door in the closed state.

  The evaluation value for a given $\bm{s}$ is the sum of the following three values (a)--(c).
  First, it is most important to correctly recognize as many states of the images in the dataset as possible.
  That is, we use the number $N^{img}_{correct}$ of images for which the Correct Rate $R^{i}_{correct}$ exceeds 0.5 for each image $i$ $(1 \leq i \leq N^{train}_{data})$ in the dataset.
  This $R^{i}_{correct}$ is the rate of correct answers for all questions where the value of $\bm{s}$ is 1, and for the image $i$ augmented by RGBShift.
  In practice, we use the ratio of the number of correctly recognized images among all images (a) $R^{img}_{correct}=N^{img}_{correct}/N^{train}_{data}$ as the evaluation value.
  Second, if $R^{img}_{correct}$ is the same, the Correct Rate should be higher.
  Therefore, as the evaluation value, we use the expected value of the Correct Rate, i.e. the average value (b) $R^{ave}_{correct}$ of $R^{i}_{correct}$ for all images in the dataset.
  For (a) and (b), we also define $R'_{correct}=N_{correct}/(N_{correct}+N_{wrong}+N_{invalid})$ and use $R^{img'}_{correct}$ and $R^{ave'}_{correct}$ as the evaluation values for comparison.
  This is expected to have the effect of treating Invalid answers as equivalent to Wrong answers, so that questions with many Invalid answers are not selected.
  Third, if $R^{img}_{correct}$ is the same, having a fewer questions reduces the computational complexity.
  On the other hand, a larger number of questions is more likely to provide stable state recognition based on a larger number of answers.
  Therefore, we use the ratio of questions used for the combination among all questions (c) $R_{q}=N^{s}_{q}/N_{q}$ (where $N^{s}_{q}$ is the number of 1 in $\bm{s}$) as the evaluation value.
  In this study, we compare the results of optimization using the following four evaluation values $E$,
  \begin{align}
    E_{+} = \alpha R^{img}_{correct}+R^{ave}_{correct}+\beta R_{q}\\
    E'_{+} = \alpha R^{img'}_{correct}+R^{ave'}_{correct}+\beta R_{q}\\
    E_{-} = \alpha R^{img}_{correct}+R^{ave}_{correct}-\beta R_{q}\\
    E'_{-} = \alpha R^{img'}_{correct}+R^{ave'}_{correct}-\beta R_{q}
  \end{align}
  where $\alpha$ and $\beta$ are weight coefficients, and in this study, $\alpha=100$ and $\beta=0.1$ to prioritize the evaluation values in the order of (a), (b) and (c).
  Likewise, we denote the question combinations obtained by using these evaluation values as $\bm{s}_{+}, \bm{s}'_{+}, \bm{s}_{-}, \bm{s}'_{-}$.

  The detailed setup of a genetic algorithm is shown below.
  The library DEAP \cite{fortin2012deap} is used, and the function cxTwoPoint is used for crossover with a probability of 50\%, and the function mutFlipBit is used for mutation with a probability of 20\%.
  Individuals are selected by the function selTournament, and the tournament size is set to 5.
  The number of individuals and generations are set to 1000 and 200, respectively.

  In addition, in this study, we prepare $\bm{s}_{\{does, is\}}$ where all the questions of the form ``Does -?'' or ``Is -?'' is used.
  We also denote $\bm{s}$ where all values of $\bm{s}$ are 1 as $\bm{s}_{all}$.
  After optimizing the question combination using the training dataset, we prepare a test dataset different from the training dataset and compare $N^{img}_{correct}$, $R^{ave}_{correct}$, $R_{invalid}$ and $N^{s}_{q}$ ($R_{invalid}$ is the ratio of Invalid responses among all responses).
  The number of images in a test dataset is the same as that in a training dataset.
}%
{%
  \secref{subsec:recognition}で述べた多数の質問文について, 最も良い精度を出すことが可能な質問の組み合わせを探索する.
  本研究ではこれに遺伝的アルゴリズムを用いた.
  質問の組み合わせを二値ベクトルである$\bm{s}$ $(\in \{0, 1\}^{N_q}$, $N_q$は質問の個数を表す$)$により表現し, これを最適化する.
  なお, 値が1である質問を用い, 0である質問は用いない.
  最適化用のデータセットとして, 本研究では$N^{train}_{data}=20$枚の画像を集めた.
  これは例えば, ドアの開閉認識であれば, ドアが開いている状態10枚, 閉まっている状態10枚である.
  このデータを用いて質問の組み合わせを最適化する.

  ある$\bm{s}$に対する評価値は以下の3つの値(1)--(3)の合計とする.
  (1) まず, 最も重要なことは, データセットとなる画像の状態をなるべく多く正しく認識することである.
  つまり, データセットに含まれるそれぞれの画像$i$ $(1 \leq i \leq N^{train}_{data})$に関する正解率$R^{i}_{correct}$が0.5を超える画像の枚数$N^{img}_{correct}$を用いる.
  この$R^{i}_{correct}$は, $\bm{s}$の値が1である全ての質問とRGBShiftによりかさ増しされた画像$i$に関する回答を合わせた際の正解率である.
  なお, 実際には, 正しく認識できた画像の枚数の割合$R^{img}_{correct}=N^{img}_{correct}/N^{train}_{data}$を用いる.
  (2) 次に, $R^{img}_{correct}$が同じであれば, より正解率が高いことが望ましい.
  そのため, 正解率の期待値, つまりデータセットに含まれる全ての画像に関する$R^{i}_{correct}$の平均値$R^{ave}_{correct}$を用いる.
  なお, (1)と(2)については, $R'_{correct}=N_{correct}/(N_{correct}+N_{wrong}+N_{invalid})$を定義し, これを用いた$R^{img'}_{correct}$と$R^{ave'}_{correct}$を評価値として利用する場合を比較する.
  これは, Invalidな答えをWrongと同等に扱うことで, Invalidな回答が多い質問を選択しなくなる効果が期待される.
  (3) 最後に, $R^{img}_{correct}$が同じであれば, より質問数が少ない方が計算量を削減することができる.
  一方で, 質問数が多いほうがより多数の意見を元に安定的に状態認識できる可能性が高い.
  よって, ここでは質問の割合$R_{q}=N^{s}_{q}/N_{q}$を用いる(ここで, $N^{s}_{q}$は$\bm{s}$における1の数を表す).
  本研究では以下の4つの評価値$E$により最適化を行った結果を比較する.
  \begin{align}
    E_{+} = \alpha R^{img}_{correct}+R^{ave}_{correct}+\beta R_{q}\\
    E'_{+} = \alpha R^{img'}_{correct}+R^{ave'}_{correct}+\beta R_{q}\\
    E_{-} = \alpha R^{img}_{correct}+R^{ave}_{correct}-\beta R_{q}\\
    E'_{-} = \alpha R^{img'}_{correct}+R^{ave'}_{correct}-\beta R_{q}
  \end{align}
  ここで, $\alpha$と$\beta$は重み付けの係数であり, 本研究では(1), (2), (3)の順で評価を優先させるために, $\alpha=100$, $\beta=0.1$としている.
  同様に, これらによって得られた質問の組み合わせを$\bm{s}_{+}, \bm{s}'_{+}, \bm{s}_{-}, \bm{s}'_{-}$と表す.

  細かい遺伝的アルゴリズムの設定について示す.
  ライブラリとしてDeap \cite{fortin2012deap}を用い, 50\%の確率で関数cxTwoPointにより交叉, 20\%の確率で関数mutFlipBitにより突然変異を行う.
  個体選択は関数selTournamentで行い, トーナメントサイズを5とする.
  個体数は1000, 世代数は200とする.

  なお, 本研究ではこの他にも, ``Does -?''または``Is -?''のそれぞれの形式の質問を全て1とした$\bm{s}_{\{does, is\}}$を用意する.
  また, 全ての値が1である$\bm{s}$を$\bm{s}_{all}$と表記する.
  訓練データを用いて質問を最適化した後, 訓練用のデータセットとは異なるテスト用データセットを用意し, それぞれの$\bm{s}$について, $N^{img}_{correct}$, $R^{ave}_{correct}$, $R_{invalid}$, $N^{s}_{q}$を比較する(なお, $R_{invalid}$は全ての回答におけるInvalidの割合を表す).
  なお, テスト用データセットは訓練用データセットと同じ枚数だけ用意している.
}%

\section{Experiments} \label{sec:experiment}
\switchlanguage%
{%
  The set of questions and representative images used in the experiments is shown in \figref{figure:experiment}.
  The experiments are conducted to recognize whether a basic door, an elevator door, a cabinet door, a refrigerator door, various doors, a microwave door, and a transparent door are open or closed, whether a light or a display is on or off, whether a handbag is open or closed, and whether water is running or not.
}%
{%
  本研究の実験で用いた画像の一部とその質問の組み合わせを\figref{figure:experiment}に示す.
  標準的なドア・エレベータのドア・棚のドア・冷蔵庫のドア・様々なドア・レンジのドア・透明なドアの開閉, 明かり・ディスプレイのオンオフ, カバンの開閉, 水が流れているかどうかを認識する実験を行う.
}%

\subsection{Door Experiment} \label{subsec:door}
\switchlanguage%
{%
  \tabref{table:door} shows the recognition results of the open/closed state of a standard door.
  Although only 15/20 of $\bm{s}_{does}$ were correctly recognized, $\bm{s}_{is}$, $\bm{s}_{all}$ and the optimization results show that all images were correctly recognized.
  While $\bm{s}_{+}$ and $\bm{s}'_{+}$ use a large number of questions, $\bm{s}_{-}$ and $\bm{s}'_{-}$ use only the best question.
  When comparing $\bm{s}_{+}$ and $\bm{s}'_{+}$, $\bm{s}'_{+}$ has less $R_{invalid}$ and uses fewer questions.
  This is considered to be because $\bm{s}'_{+}$ treats Invalid as the same as Wrong, so the choices are limited to those that are less likely to be Invalid.
}%
{%
  \tabref{table:door}に標準的なドアの開閉状態の認識結果を示す.
  $\bm{s}_{does}$は15/20しか正確に認識できなかったものの, $\bm{s}_{is}$や$\bm{s}_{all}$, 最適化結果は全ての画像を正しく認識出来ていることがわかる.
  $\bm{s}_{+}$や$\bm{s}'_{+}$は質問の数が多いが, $\bm{s}_{-}$や$\bm{s}'_{-}$は質問の数が1と, 最も良い質問だけを用いている.
  $\bm{s}_{+}$と$\bm{s}'_{+}$を比べると, $\bm{s}'_{+}$の方が$R_{invalid}$が少なく, 質問の数も少ない.
  これは, $\bm{s}'_{+}$がInvalidをWrongと同等に扱っているため, Invalidが少ないような選択肢に絞られていることが理由だと考えられる.
}%

\begin{table}[htb]
  \centering
  \caption{The result of door experiment}
  \begin{tabular}{l|cccc}
    & $N^{img}_{correct}$ & $R^{ave}_{correct}$ & $R_{invalid}$ & $N^{s}_{q}$ \\ \hline
    $\bm{s}_{+}$ & 20 / 20 & 0.981 & 0.035 & 11 / 16 \\
    $\bm{s}'_{+}$ & 20 / 20 & 0.977 & 0.001 & 8 / 16 \\
    $\bm{s}_{-}, \bm{s}'_{-}$ & 20 / 20 & 1.000 & 0.000 & 1 / 16 \\
    $\bm{s}_{does}$ & 15 / 20 & 0.800 & 0.000 & 8 / 16 \\
    $\bm{s}_{is}$ & 20 / 20 & 0.904 & 0.056 & 8 / 16 \\
    $\bm{s}_{all}$ & 20 / 20 & 0.852 & 0.028 & 16 / 16
  \end{tabular}
  \label{table:door}
\end{table}

\subsection{Elevator Experiment} \label{subsec:elevator}
\switchlanguage%
{%
  \tabref{table:elevator} shows the recognition results of the open/closed state of an elevator door.
  Here, two kinds of words, ``door'' and ``elevator door'', are used as the wording.
  While $\bm{s}_{\{does, is, all\}}$ makes recognition mistakes for some images, $\bm{s}_{\{+, -\}}$ and $\bm{s}'_{\{\{+, -\}}$ could recognize all images correctly.
  Moreover, the optimization results of $\bm{s}_{\{+, -\}}$ and $\bm{s}'_{\{+, -\}}$ are all consistent, suggesting that using only the best question is significantly more accurate than using the other questions.
  The best question was ``Does this image look like the door is open?''.
  In addition, when comparing $\bm{s}_{does}$ and $\bm{s}_{is}$, $\bm{s}_{does}$ has lower $R^{ave}_{correct}$ and smaller $R_{invalid}$ than $\bm{s}_{is}$, which is the same tendency as in \secref{subsec:door}.
}%
{%
  \tabref{table:elevator}にエレベータドアの開閉状態の認識結果を示す.
  ここでは, 言葉遣いとして, ``door''と``elevator door'の2種類の単語を用いている.
  $\bm{s}_{\{does, is, all\}}$はいくつかの画像について認識ミスをしているが, $\bm{s}_{\{+, -\}}$や$\bm{s}'_{\{+, -\}}$はどれも全ての画像を正しく認識できている.
  また, $\bm{s}_{\{+, -\}}$, $\bm{s}'_{\{+, -\}}$の最適化結果は全て一致しており, 一つの質問だけを用いる場合が他の質問を用いる場合に比べて圧倒的に精度が良かったと考えられる.
  なお, その質問は``Does this image look like the door is open?''であった.
  加えて, \secref{subsec:door}にも同じ傾向が見られるが, $\bm{s}_{does}$と$\bm{s}_{is}$を比べると, $\bm{s}_{does}$の方が$R^{ave}_{correct}$が低く, $R_{invalid}$が小さい.
}%

\begin{table}[htb]
  \centering
  \caption{The result of elevator experiment}
  \begin{tabular}{l|cccc}
    & $N^{img}_{correct}$ & $R^{ave}_{correct}$ & $R_{invalid}$ & $N^{s}_{q}$ \\ \hline
    $\bm{s}_{\{+, -\}}, \bm{s}'_{\{+, -\}}$ & 20 / 20 & 0.992 & 0.000 & 1 / 32 \\
    $\bm{s}_{does}$ & 17 / 20 & 0.750 & 0.000 & 16 / 32 \\
    $\bm{s}_{is}$ & 17 / 20 & 0.796 & 0.048 & 16 / 32 \\
    $\bm{s}_{all}$ & 18 / 20 & 0.773 & 0.024 & 32 / 32
  \end{tabular}
  \label{table:elevator}
\end{table}

\subsection{Cabinet Experiment}
\switchlanguage%
{%
  \tabref{table:cabinet} shows the recognition results of the open/closed state of a cabinet door.
  Here, four words, ``door'', ``cabinet door'', ``kitchen cabinet door'', and ``shelf door'', are used as the wording.
  For all question combinations, all states are successfully recognized.
  The characteristics are the same as above experiments, $\bm{s}_{+}$ has larger $R_{invalid}$ and $N^{s}_{q}$ than $\bm{s}'_{+}$.
  Also, $\bm{s}_{does}$ has lower $R^{ave}_{correct}$ and smaller $R_{invalid}$ than $\bm{s}_{is}$.
}%
{%
  \tabref{table:cabinet}に棚のドアの開閉状態の認識結果を示す.
  ここでは, 言葉遣いとして, ``door'', ``cabinet door'', ``kitchen cabinet door'', ``shelf door''の4種類の単語を用いている.
  全ての質問の組み合わせについて, 全ての状態認識に成功している.
  その特性はこれまで同様で, $\bm{s}_{+}$の方が$\bm{s}'_{+}$に比べて$R_{invalid}$が大きく, $N^{s}_{q}$も多い.
  また, $\bm{s}_{does}$の方が$\bm{s}_{is}$に比べて, $R^{ave}_{correct}$が低く, $R_{invalid}$が小さい.
}%

\begin{table}[htb]
  \centering
  \caption{The result of cabinet experiment}
  \begin{tabular}{l|cccc}
    & $N^{img}_{correct}$ & $R^{ave}_{correct}$ & $R_{invalid}$ & $N^{s}_{q}$ \\ \hline
    $\bm{s}_{+}$ & 20 / 20 & 0.963 & 0.082 & 26 / 64 \\
    $\bm{s}'_{+}$ & 20 / 20 & 0.971 & 0.000 & 17 / 64 \\
    $\bm{s}_{-}, \bm{s}'_{-}$ & 20 / 20 & 1.000 & 0.000 & 1 / 64 \\
    $\bm{s}_{does}$ & 20 / 20 & 0.865 & 0.000 & 32 / 64 \\
    $\bm{s}_{is}$ & 20 / 20 & 0.878 & 0.125 & 32 / 64 \\
    $\bm{s}_{all}$ & 20 / 20 & 0.873 & 0.062 & 64 / 64
  \end{tabular}
  \label{table:cabinet}
\end{table}

\subsection{Refrigerator Experiment}
\switchlanguage%
{%
  \tabref{table:refrigerator} shows the recognition results of the open/closed state of a refrigerator door.
  The optimized question combinations outperform $\bm{s}_{\{does, is, all\}}$ in accuracy.
  Regarding all optimization results, $N^{img}_{correct}$ is 20/20 for the training dataset used in the optimization, while it is 19/20 in some cases for the test dataset.
  Other characteristics are similar to those in the above experiments.
}%
{%
  \tabref{table:refrigerator}に冷蔵庫のドアの開閉状態の認識結果を示す.
  $\bm{s}_{\{does, is, all\}}$に比べて最適化された質問の組み合わせが精度で上回っている.
  全ての最適化結果について, 最適化時に使用した訓練データセットでは$N^{img}_{correct}$は20/20であったが, テストデータセットでは一部19/20となる場合がった.
  その他の特性はこれまでの実験と大差ない.
}%

\begin{table}[htb]
  \centering
  \caption{The result of refrigerator experiment}
  \begin{tabular}{l|cccc}
    & $N^{img}_{correct}$ & $R^{ave}_{correct}$ & $R_{invalid}$ & $N^{s}_{q}$ \\ \hline
    $\bm{s}_{+}, \bm{s}'_{+}$ & 19 / 20 & 0.964 & 0.000 & 3 / 32 \\
    $\bm{s}_{-}$ & 19 / 20 & 0.975 & 0.000 & 1 / 32 \\
    $\bm{s}'_{-}$ & 20 / 20 & 0.983 & 0.000 & 1 / 32 \\
    $\bm{s}_{does}$ & 15 / 20 & 0.727 & 0.000 & 16 / 32 \\
    $\bm{s}_{is}$ & 17 / 20 & 0.737 & 0.025 & 16 / 32 \\
    $\bm{s}_{all}$ & 17 / 20 & 0.732 & 0.013 & 32 / 32
  \end{tabular}
  \label{table:refrigerator}
\end{table}

\subsection{Various Doors Experiment}
\switchlanguage%
{%
  \tabref{table:various} shows the recognition results of the open/closed states of all the four doors described so far.
  Only ``door'' is used as the wording, but the dataset given includes the standard door, the elevator door, the cabinet door, and the refrigerator door.
  80 door images were used for the training and test datasets, respectively.
  The optimized question combination outperforms $\bm{s}_{\{does, is, all\}}$ in accuracy.
  In other words, if the state to be recognized has similar characteristics, the state recognition is possible without changing the questions, and there is no need to construct each recognizer individually.
  Also, the number of images that can be recognized with only one question is 72/80, which is lower than 76/80 for optimization results, indicating that it is possible to achieve better recognition accuracy by using appropriate question combinations.
}%
{%
  \tabref{table:various}にこれまで述べた4つのドア全ての開閉状態の認識結果を示す.
  wordingとしては``door''のみ用いるが, 与えるデータセットが, これまでの標準的なドア, エレベータのドア, 棚のドア, 冷蔵庫のドアで, 訓練時テスト時それぞれ80枚ずつ存在する.
  $\bm{s}_{\{does, is, all\}}$に比べて最適化された質問の組み合わせが精度で上回っている.
  つまり, 似た特性の状態認識であれば, 特に質問を変化させなくとも状態認識が可能であり, 一つ一つ個別にプログラミングする必要がない.
  また, 一つの質問だけで認識可能な画像数は72/80であったため, 適切な質問の組み合わせを利用することで, より良い認識精度を出すことが可能であることもわかる.
}%

\begin{table}[htb]
  \centering
  \caption{The result of various doors experiment}
  \begin{tabular}{l|cccc}
    & $N^{img}_{correct}$ & $R^{ave}_{correct}$ & $R_{invalid}$ & $N^{s}_{q}$ \\ \hline
    $\bm{s}_{+}$ & 76 / 80 & 0.884 & 0.075 & 7 / 16 \\
    $\bm{s}_{-}$ & 76 / 80 & 0.890 & 0.041 & 3 / 16 \\
    $\bm{s}'_{\{+, -\}}$ & 76 / 80 & 0.890 & 0.004 & 4 / 16 \\
    $\bm{s}_{does}$ & 64 / 80 & 0.789 & 0.000 & 8 / 16 \\
    $\bm{s}_{is}$ & 73 / 80 & 0.857 & 0.092 & 8 / 16 \\
    $\bm{s}_{all}$ & 72 / 80 & 0.823 & 0.046 & 16 / 16
  \end{tabular}
  \label{table:various}
\end{table}

\subsection{Microwave Experiment}
\switchlanguage%
{%
  \tabref{table:microwave} shows the recognition results of the open/closed state of a microwave door.
  Here, four words, ``door'', ``microwave door'', ``oven door'', and ``microwave oven door'', are used as the wording.
  The optimized $\bm{s}_{\{+, -\}}$ in particular outperforms $\bm{s}_{\{does, is, all\}}$ in accuracy.
  On the other hand, $N^{img}_{correct}$ for $\bm{s}'_{\{+, -\}}$ is smaller than that for $\bm{s}_{\{+, -\}}$.
  This is considered to be because $\bm{s}'_{\{+, -\}}$ treats Invalid responses as Wrong, and thus questions with many Invalid responses but with accurate state recognition are lost as choices.
  The value of $R_{invalid}$ for $\bm{s}_{\{+, -\}}$ is larger than that for $\bm{s}_{\{does, is, all\}}$, indicating that $\bm{s}_{\{+, -\}}$ actively uses questions that output Invalid responses.
  Note that the microwave door is not included in the various doors experiment because the simple wording of ``door'' makes it difficult to recognize the state of the microwave door.
}%
{%
  \tabref{table:microwave}にレンジのドアの開閉状態の認識結果を示す.
  ここでは, 言葉遣いとして, ``door'', ``microwave door'', ``oven door'', ``microwave open door''の4種類の単語を用いている.
  $\bm{s}_{\{does, is, all\}}$に比べて, 特に最適化された$\bm{s}_{\{+, -\}}$は精度で上回っている.
  一方, $\bm{s}'_{\{+, -\}}$は$\bm{s}_{\{+, -\}}$に比べて$N^{img}_{correct}$が小さい.
  これは, $\bm{s}'_{\{+, -\}}$がInvalidな回答をWrongと扱うため, Invalidな回答は多いが正確な状態認識が可能な質問を選択肢として失ってしまっているためだと考えられる.
  $\bm{s}_{\{+, -\}}$は$R_{invalid}$の値が$\bm{s}_{\{does, is, all\}}$に比べても大きく, 積極的に無効な選択肢を出力する質問群を利用していることが分かる.
  なお, 単純な``door''だと精度が低く, various door実験には含めていない.
}%

\begin{table}[htb]
  \centering
  \caption{The result of microwave experiment}
  \begin{tabular}{l|cccc}
    & $N^{img}_{correct}$ & $R^{ave}_{correct}$ & $R_{invalid}$ & $N^{s}_{q}$ \\ \hline
    $\bm{s}_{+}$ & 17 / 20 & 0.703 & 0.315 & 6 / 64 \\
    $\bm{s}_{-}$ & 16 / 20 & 0.689 & 0.210 & 5 / 64 \\
    $\bm{s}'_{+}$ & 12 / 20 & 0.546 & 0.008 & 47 / 64 \\
    $\bm{s}'_{-}$ & 12 / 20 & 0.596 & 0.015 & 8 / 64 \\
    $\bm{s}_{does}$ & 12 / 20 & 0.506 & 0.000 & 32 / 64 \\
    $\bm{s}_{is}$ & 15 / 20 & 0.621 & 0.084 & 32 / 64 \\
    $\bm{s}_{all}$ & 14 / 20 & 0.561 & 0.042 & 64 / 64
  \end{tabular}
  \label{table:microwave}
\end{table}

\subsection{Transparent Door Experiment}
\switchlanguage%
{%
  \tabref{table:transparent} shows the recognition results of the open/closed state of a transparent door.
  Here, four words, ``door'', ``transparent door'', ``glass door'', and ``window'', are used as the wording.
  The optimized question combination significantly outperforms $\bm{s}_{\{does, is, all\}}$ in accuracy.
  The accuracy of $\bm{s}_{\{does, is, all\}}$ is 10/20 for $N^{img}_{correct}$, and this is due to the fact that the door is transparent, which leads to the incorrect recognition that the door is always open.
  On the other hand, some questions can correctly recognize the state even if the door is transparent, such as when asking the two questions ``Does this image look like a window is closed?'' and ``Is this window open?'' used for $\bm{s}'_{-}$.
  The questions obtained from other optimization results also suggest that the wording of ``window'' is very important.
}%
{%
  \tabref{table:transparent}に透明なドアの開閉状態の認識結果を示す.
  ここでは, 言葉遣いとして, ``door'', ``transparent door'', ``glass door'', ``window''の4種類の単語を用いている.
  $\bm{s}_{\{does, is, all\}}$に比べて最適化された質問の組み合わせが精度で大きく上回っている.
  $\bm{s}_{\{does, is, all\}}$は$N^{img}_{correct}$が10/20であるが, これはドアが透明であるため, 常にドアが開いていると勘違いしてしまっていることに起因する.
  一方で, 一部の質問はドアが透明でも正しく状態認識可能であり, 例えば, $\bm{s}'_{-}$に用いられた質問は, ``Does this image look like a window is closed?''と``Is this window open?''の2つである.
  他の最適化結果で得られた質問からも, ``window''というwordingが非常に重要であることが示唆された.
}%

\begin{table}[htb]
  \centering
  \caption{The result of transparent door experiment}
  \begin{tabular}{l|cccc}
    & $N^{img}_{correct}$ & $R^{ave}_{correct}$ & $R_{invalid}$ & $N^{s}_{q}$ \\ \hline
    $\bm{s}_{\{+, -\}}, \bm{s}'_{+}$ & 15 / 20 & 0.692 & 0.000 & 5 / 64 \\
    $\bm{s}'_{-}$ & 16 / 20 & 0.729 & 0.000 & 2 / 64 \\
    $\bm{s}_{does}$ & 10 / 20 & 0.574 & 0.000 & 32 / 64 \\
    $\bm{s}_{is}$ & 10 / 20 & 0.465 & 0.116 & 32 / 64 \\
    $\bm{s}_{all}$ & 10 / 20 & 0.519 & 0.058 & 64 / 64
  \end{tabular}
  \label{table:transparent}
\end{table}


\subsection{Light Experiment} \label{subsec:light}
\switchlanguage%
{%
  Next, we will show that it is possible to recognize not only the open/closed state of the door as we have seen so far, but also various other states.
  \tabref{table:light} shows the recognition results of the on/off state of a room light.
  Here, ``light'' and ``electricity'' are used as the wording.
  The optimized question combination significantly outperforms $\bm{s}_{\{does, is, all\}}$ in accuracy.
  $\bm{s}_{\{+, -\}}$ outperforms $\bm{s}'_{\{+, -\}}$ in $N^{img}_{correct}$, although $R^{ave}_{correct}$ is smaller.
  This is considered to be because $\bm{s}'_{\{+, -\}}$ treats Invalid responses as Wrong, and thus questions with many Invalid responses but with accurate state recognition are lost as choices.
  Note that the word used in all the optimized question combinations is ``electricity''.
  Other characteristics are the same as in the above experiments.
}%
{%
  これまで見てきたドアの開閉状態だけではなく, その他多様な状態認識が可能であることを示していく.
  \tabref{table:light}に部屋の明かりのオンオフ状態の認識結果を示す.
  ここでは, 言葉遣いとして, ``light''と``electricity''の2種類の単語を用いている.
  $\bm{s}_{\{does, is, all\}}$に比べて最適化された質問の組み合わせが精度で大きく上回っている.
  $\bm{s}_{\{+, -\}}$は$\bm{s}'_{\{+, -\}}$に比べて$R^{ave}_{correct}$は小さいものの, $N^{img}_{correct}$は上回っている.
  これは, $\bm{s}'_{\{+, -\}}$がInvalidな回答をWrongと扱うため, Invalidな回答は多いが正確な状態認識が可能な質問を選択肢として失ってしまっているためだと考えられる.
  なお, 最適化で得られた質問に使われている単語は全て``electricity''であった.
  その他の特性はこれまでと変わらない.
}%

\begin{table}[htb]
  \centering
  \caption{The result of light experiment}
  \begin{tabular}{l|cccc}
    & $N^{img}_{correct}$ & $R^{ave}_{correct}$ & $R_{invalid}$ & $N^{s}_{q}$ \\ \hline
    $\bm{s}_{\{+, -\}}$ & 20 / 20 & 0.799 & 0.094 & 3 / 32 \\
    $\bm{s}'_{\{+, -\}}$ & 17 / 20 & 0.856 & 0.000 & 3 / 32 \\
    $\bm{s}_{does}$ & 10 / 20 & 0.598 & 0.000 & 16 / 32 \\
    $\bm{s}_{is}$ & 10 / 20 & 0.573 & 0.069 & 16 / 32 \\
    $\bm{s}_{all}$ & 11 / 20 & 0.586 & 0.034 & 32 / 32
  \end{tabular}
  \label{table:light}
\end{table}

\subsection{Display Experiment}
\switchlanguage%
{%
  \tabref{table:display} shows the recognition results of the on/off state of a display.
  Here, four words, ``display'', ``monitor'', ``computer monitor'', and ``TV'', are used as the wording.
  The optimized question combination significantly outperforms $\bm{s}_{\{does, is, all\}}$ in accuracy.
  $\bm{s}_{-}$ and $\bm{s}'_{-}$ are identical, and only one question ``Does this image look like this display is off?'' is used.
  On the other hand, the combination of many questions like $\bm{s}_{+}$ or $\bm{s}'_{+}$ is somewhat more accurate than $\bm{s}_{-}$ or $\bm{s}'_{-}$ with $N^{img}_{correct}$ of 20/20.
}%
{%
  \tabref{table:display}にディスプレイのオンオフ状態の認識結果を示す.
  ここでは, 言葉遣いとして, ``display'', ``monitor'', ``computer monitor'', ``TV''の4種類の単語を用いている.
  $\bm{s}_{\{does, is, all\}}$に比べて最適化された質問の組み合わせが精度で大きく上回っている.
  $\bm{s}_{-}$と$\bm{s}'_{-}$は同一で, 用いられた質問は``Does this image look like this display is off?''一つだけであった.
  一方, $\bm{s}_{+}$や$\bm{s}'_{+}$のように多数の質問を組み合わせた場合の方が$N^{img}_{correct}$が20/20で多少精度が高い.
}%

\begin{table}[htb]
  \centering
  \caption{The result of display experiment}
  \begin{tabular}{l|cccc}
    & $N^{img}_{correct}$ & $R^{ave}_{correct}$ & $R_{invalid}$ & $N^{s}_{q}$ \\ \hline
    $\bm{s}_{+}$ & 20 / 20 & 0.950 & 0.557 & 20 / 64 \\
    $\bm{s}'_{+}$ & 20 / 20 & 0.957 & 0.000 & 6 / 64 \\
    $\bm{s}_{-}, \bm{s}'_{-}$ & 19 / 20 & 0.925 & 0.000 & 1 / 64 \\
    $\bm{s}_{does}$ & 12 / 20 & 0.716 & 0.000 & 32 / 64 \\
    $\bm{s}_{is}$ & 14 / 20 & 0.693 & 0.491 & 32 / 64 \\
    $\bm{s}_{all}$ & 14 / 20 & 0.705 & 0.245 & 64 / 64
  \end{tabular}
  \label{table:display}
\end{table}

\subsection{Handbag Experiment} \label{subsec:handbag}
\switchlanguage%
{%
  \tabref{table:handbag} shows the recognition results of the open/closed state of a handbag.
  Here, ``bag'' and ``handbag'' are used as the wording.
  Compared to $\bm{s}_{\{does, is, all\}}$, the optimized question combination is much more accurate.
  Other characteristics are the same as in previous experiments.
}%
{%
  \tabref{table:bag}にバッグの開閉態の認識結果を示す.
  ここでは, 言葉遣いとして, ``bag''と``handbag''の2種類の単語を用いている.
  $\bm{s}_{\{does, is, all\}}$に比べて最適化された質問の組み合わせが精度で大きく上回っている.
  その他の特性はこれまでと変わらない.
}%

\begin{table}[htb]
  \centering
  \caption{The result of handbag experiment}
  \begin{tabular}{l|cccc}
    & $N^{img}_{correct}$ & $R^{ave}_{correct}$ & $R_{invalid}$ & $N^{s}_{q}$ \\ \hline
    $\bm{s}_{+}$ & 18 / 20 & 0.815 & 0.497 & 8 / 32 \\
    $\bm{s}_{-}$ & 18 / 20 & 0.815 & 0.195 & 5 / 32 \\
    $\bm{s}'_{\{+, -\}}$ & 18 / 20 & 0.750 & 0.000 & 3 / 32 \\
    $\bm{s}_{does}$ & 15 / 20 & 0.670 & 0.002 & 16 / 32 \\
    $\bm{s}_{is}$ & 11 / 20 & 0.530 & 0.461 & 16 / 32 \\
    $\bm{s}_{all}$ & 14 / 20 & 0.622 & 0.232 & 32 / 32
  \end{tabular}
  \label{table:handbag}
\end{table}

\subsection{Water Experiment}
\switchlanguage%
{%
  \tabref{table:water} shows the recognition results of whether water is running or not.
  Here, after ``water is running'', two modifications are added: ``in {a, the, this, that} sink'' or ``from {a, the, this, that} faucet''.
  The optimized question combination significantly outperforms $\bm{s}_{\{does, is, all\}}$ in accuracy.
  Similarly to \secref{subsec:light}, $\bm{s}_{\{+, -\}}$ outperforms $\bm{s}'_{\{+, -\}}$ in $N^{img}_{correct}$, indicating the importance of questions with many Invalid responses.
  Note that $N^{img}_{correct}$ is 19/20 for all the data used in the optimization, indicating that the accuracy is somewhat lower for the test dataset.
  Also, all the questions used in the optimization results were the ones with the modification ``from {a, the, this, that} faucet''.
}%
{%
  \figref{table:water}に水が出ているかどうかの状態の認識結果を示す.
  ここでは, ``water is running''の後に, ``in {a, the, this, that} sink''または``from {a, the, this, that} faucet''の2種類の修飾を加えている.
  $\bm{s}_{\{does, is, all\}}$に比べて最適化された質問の組み合わせが精度で大きく上回っている.
  \secref{subsec:light}と同様に, $\bm{s}_{\{+, -\}}$は$\bm{s}'_{\{+, -\}}$に比べて$N^{img}_{correct}$が上回っており, Invalidな回答が多い質問の重要性が伺える.
  なお, 最適化時に用いたデータセットについてはどれも$N^{img}_{correct}$が19/20であったため, 多少テストデータセットの際には精度が落ちている.
  また, 最適化結果で用いられた質問は全て``from {a, the, this, that} faucet''の修飾が加わったものであった.
}%

\begin{table}[htb]
  \centering
  \caption{The result of water experiment}
  \begin{tabular}{l|cccc}
    & $N^{img}_{correct}$ & $R^{ave}_{correct}$ & $R_{invalid}$ & $N^{s}_{q}$ \\ \hline
    $\bm{s}_{+}$ & 17 / 20 & 0.797 & 0.179 & 6 / 32 \\
    $\bm{s}_{-}$ & 17 / 20 & 0.796 & 0.033 & 5 / 32 \\
    $\bm{s}'_{\{+, -\}}$ & 15 / 20 & 0.735 & 0.033 & 5 / 32 \\
    $\bm{s}_{does}$ & 12 / 20 & 0.580 & 0.000 & 16 / 32 \\
    $\bm{s}_{is}$ & 13 / 20 & 0.667 & 0.114 & 16 / 32 \\
    $\bm{s}_{all}$ & 12 / 20 & 0.622 & 0.057 & 32 / 32
  \end{tabular}
  \label{table:water}
\end{table}

\section{Discussion} \label{sec:discussion}
\switchlanguage%
{%
  The following is a summary of the important points from the obtained experimental results.
  \begin{itemize}
    \item The accuracy of using many questions such as $\bm{s}_{\{does, is, all\}}$ at once is not high, and it is necessary to use only appropriate questions.
    \item The easier the state recognition is, the higher the accuracy is even if there is only one question.
    \item In many cases, one question alone is not accurate enough, and it is possible to construct a recognizer with better accuracy by appropriately combining multiple questions.
    \item Invalid can be treated the same as Wrong to reduce the number of Invalid answers, but this does not necessarily lead to better accuracy. Rather, some important questions are often left out of the choices and the accuracy is often decreased.
    \item In some state recognition, having a large number of questions can achieve higher accuracy than having fewer questions, even for a dataset different from the one used in the optimization.
  \end{itemize}
  Taking all of the above into consideration, we should basically use the questions obtained by $\bm{s}_{\{+, -\}}$.
  $\bm{s}_{+}$ is recommended for accurate recognition even if it takes longer, while $\bm{s}_{-}$ is recommended for accurate recognition in a shorter time span.

  Some other notable properties are:
  \begin{itemize}
    \item There are more Invalid responses for questions in $\bm{s}_{is}$ than for questions in $\bm{s}_{does}$.
    \item The recognition accuracy is better for questions in $\bm{s}_{is}$ than for questions in $\bm{s}_{does}$.
    \item All the questions obtained by optimization use the same wording.
  \end{itemize}
  These characteristics are considered to vary depending on the model used for VQA, and are not universally applicable knowledge.
  However, by using this study, it is possible to obtain an appropriate combination of questions without worrying about the characteristics that vary from model to model, by simply providing some dataset.

  The important contribution of this study is that the recognizer can be constructed intuitively by the spoken language, without the need for retraining of the network or individual programming.
  In addition, only a single pre-trained model is required, and the resource is very easy to manage.
  On the other hand, since the choice of the question sentences is heuristic, we have added an optimization mechanism to solve this problem.
  This makes it possible to construct a state recognizer very easily, in which the only part that humans have to consider is to increase the variety of questions.
  In the future, it is necessary to incorporate this study into actual tasks and to identify problems that may arise in the real world.
  In particular, we would like to examine state recognition in more cluttered situations, and recognition performance when various other objects, such as humans and robot bodies, appear in the image.
  In addition, we believe that anomaly detection and object detection will become possible in the same way as in this study, and we would like to deal with a wider range of state recognition in the near future.
}%
{%
  得られた実験結果から重要な事項を以下にまとめる.
  \begin{itemize}
    \item $\bm{s}_{\{does, is, all\}}$のような質問をまとめて利用した際の精度は高くなく, 適切な質問だけを利用する必要がある.
    \item 簡単な認識であればあるほど, 質問数は1でも問題ない.
    \item 一つの質問だけでは得意不得意が有る場合が多く, 複数の質問を適切に組み合わせることでより良い精度の認識器を構成可能である.
    \item InvalidもWrongと同等に扱うことでInvalidな回答を減らすことができるが, これは必ずしも良い精度に繋がるわけではなく, むしろ一部の重要な質問が選択肢から外れてしまい精度が下がる場合が多い.
    \item 一部の状態認識では, 質問数が多いことで少ない場合に比べて, 最適化時とは異なるデータセットに対しても高い精度を出すことが可能である.
  \end{itemize}
  これらを総合すると, 基本的に$\bm{s}_{\{+, -\}}$で得られた質問を利用すべきである.
  時間はかかっても正確な認識を目指すのであれば$\bm{s}_{+}$, 短時間でなるべく正確な認識を目指すのであれば$\bm{s}_{-}$が推奨される.

  また, この他にもいくつかの性質が見て取れた.
  \begin{itemize}
    \item $\bm{s}_{does}$に比べて$\bm{s}_{is}$の質問の方がInvalidが多い.
    \item $\bm{s}_{does}$に比べて$\bm{s}_{is}$の質問の方が認識精度が良い.
    \item 最適化によって得られた質問は皆同じwordingを用いている.
  \end{itemize}
  一方で, これらの特性はVQAに用いるモデルによっても変化すると考えられ, 普遍的に用いることが可能な知識ではない.
  しかし, 本研究を用いることで, モデルごとに異なる特性等を気にせずとも, いくつかのデータを与えるのみで, 再学習なしに適切な質問の組み合わせを得ることが可能である.

  本研究の特徴は, ネットワークの再学習や個別のプログラミング等が必要なく, 言語によって直感的に認識器が構築できる点にある.
  また, モデルも単一の大規模モデルを使用すれば良く, 管理等も非常に簡単である.
  一方で質問となる言語の選び方がヒューリスティックになってしまうため, これを解決する最適化の機構を加えた.
  これにより, 人間が考える唯一の部分は質問のバラエティを増やす部分となり, 非常に簡易に認識器が構築可能である.
  今後の展望として, この研究をさらに実際のタスクに組み込み, その際の問題点を洗い出す必要があると考えている.
  特に, よりclutteredな状況における状態認識, 人やロボット身体等, 他の多様な物体が画像中に写ってしまう場合の認識性能なども検証していきたい.
  また, 本研究と同様の形で, 異常検知や物体検出, 物体の位置関係把握等も可能になると考えており, より広範囲の状態認識も扱っていきたい.
}%

\section{CONCLUSION} \label{sec:conclusion}
\switchlanguage%
{%
  In this study, we proposed a method of binary state recognition for robots using Visual Question Answering (VQA) in a Pre-Trained Vision-Language Model (PTVLM).
  In VQA, for multiple augmented images, we integrate the answers from various questions with different forms, articles, state expressions, and wording.
  Since there are states that are easy to recognize and states that are difficult to recognition for each question, by using an appropriate combination of questions, it is possible to construct a recognizer that can accurately recognize any state.
  By optimizing this combination with a genetic algorithm, it is now possible to recognize the states of transparent doors, water, etc. that have been difficult to recognize so far.
  We believe that this method will revolutionize the recognition strategies of robots, since it does not require any retraining of the network or programming, and a recognizer of complex states can be easily constructed by simply providing a set of questions to a single model.
}%
{%
  本研究では, 事前学習済み視覚-言語モデルにおけるVisual Question Answering (VQA)を用いたロボットの二値状態認識器の構成を行った.
  VQAにおいて, ランダマイズした複数の画像と, 形式・冠詞・状態表現・言葉遣いを変化させた様々な質問による結果を統合する.
  この際, それぞれの質問には認識の得意な状態と苦手な状態が存在するため, 適切な質問の組み合わせを用いることで, どのような状態も正確に認識可能な認識器を構築することができるようになる.
  この組み合わせを遺伝的アルゴリズムで最適化することで, これまで難しかった透明なドアや水の状態認識等まで可能となった.
  ネットワークの再学習やプログラミング等が必要なく, 単一のモデルに質問セットを与えるのみで簡易に複雑な認識器を構成でき, 今後ロボットの認識行動戦略を大きく革新すると考える.
}%

{
  \bibliographystyle{IEEEtran}
  \bibliography{main}
}

\end{document}